\documentclass[sigconf]{acmart}
\usepackage{booktabs}   
\usepackage{pifont}     
\usepackage{array}      
\usepackage{makecell}  
\usepackage{amsmath}
\usepackage{listings}
\usepackage{xcolor}
\lstdefinestyle{promptstyle}{
  basicstyle=\ttfamily\footnotesize,
  frame=single,
  breaklines=true,
  columns=fullflexible,
  keepspaces=true,
  showstringspaces=false,
  captionpos=b,
  numbers=left,
  numberstyle=\tiny,
  numbersep=6pt,
  xleftmargin=1.5em,
  framexleftmargin=1.5em
}

\AtBeginDocument{%
  }

\begin{document}

\title{Do VLMs Truly "Read" Candlesticks? A Multi-Scale Benchmark for Visual Stock Price Forecasting}

\author{Kaiqi Hu}
\affiliation{%
  \institution{University of Leeds}
  \city{Leeds}
  \country{UK}
}
\email{pbtp2855@leeds.ac.uk}

\author{Linda Xiao}
\affiliation{%
  \institution{Sun Yat-sen University}
  \city{Zhuhai}
  \country{China}
}
\email{xiaold3@mail2.sysu.edu.cn}

\author{Shiyue Xu}
\affiliation{%
  \institution{Sun Yat-sen University}
  \city{Zhuhai}
  \country{China}
}
\email{xushy66@mail2.sysu.edu.cn}

\author{Ziyi Tang}
\affiliation{%
  \institution{Sun Yat-sen University}
  \city{GuangZhou}
  \country{China}
}
\email{tangzy27@mail2.sysu.edu.cn}

\author{Mingwen Liu}
\affiliation{%
  \institution{Likelihood Lab}
  \city{Guangzhou}
  \country{China}
}
\email{maxwell@alphafuture.cn}

\renewcommand{\shortauthors}{Hu et al.}

\begin{abstract}
 Vision–language models (VLMs) are increasingly applied to visual stock price forecasting, yet existing benchmarks inadequately evaluate their understanding of stock price in candlestick charts. First, prior studies fail to isolate VLMs' comprehension of visual stock price inputs, and it remains unclear whether visual inputs genuinely improve predictive performance and whether VLMs truly comprehend candlestick patterns. Further, most existing datasets and evaluation setups are designed around single-period or tabular inputs. However, human analysts strongly rely on multi-scale candlestick charts, where longer-term horizons capture trend direction and shorter-term horizons provide cues for inflection points, making it difficult to systematically assess VLMs' ability to integrate short-term and long-term visual market dynamics. To bridge this gap, we construct a multi-scale candlestick chart dataset and a standardized evaluation framework to assess VLMs’ ability to utilize multi-scale visual market signals. Evaluation combines confusion-matrix-based diagnostics with information coefficient (IC) time series metrics and includes XGBoost as a feature-based temporal baseline. Using this dataset, we benchmark representative VLMs and analyze their ability to leverage multi-scale stock price data. Experimental results show that most VLMs perform well only under persistent uptrend or downtrend conditions, while exhibiting weak predictive capability in more common market scenarios. We also identify significant prediction biases and limited sensitivity to explicitly specified forecast horizons in prompts, indicating inherent limitations in precise temporal reasoning.

\end{abstract}




\maketitle
\begin{figure*}[t]
    \centering
    \vspace{-10pt}
    \includegraphics[width=0.85\linewidth]{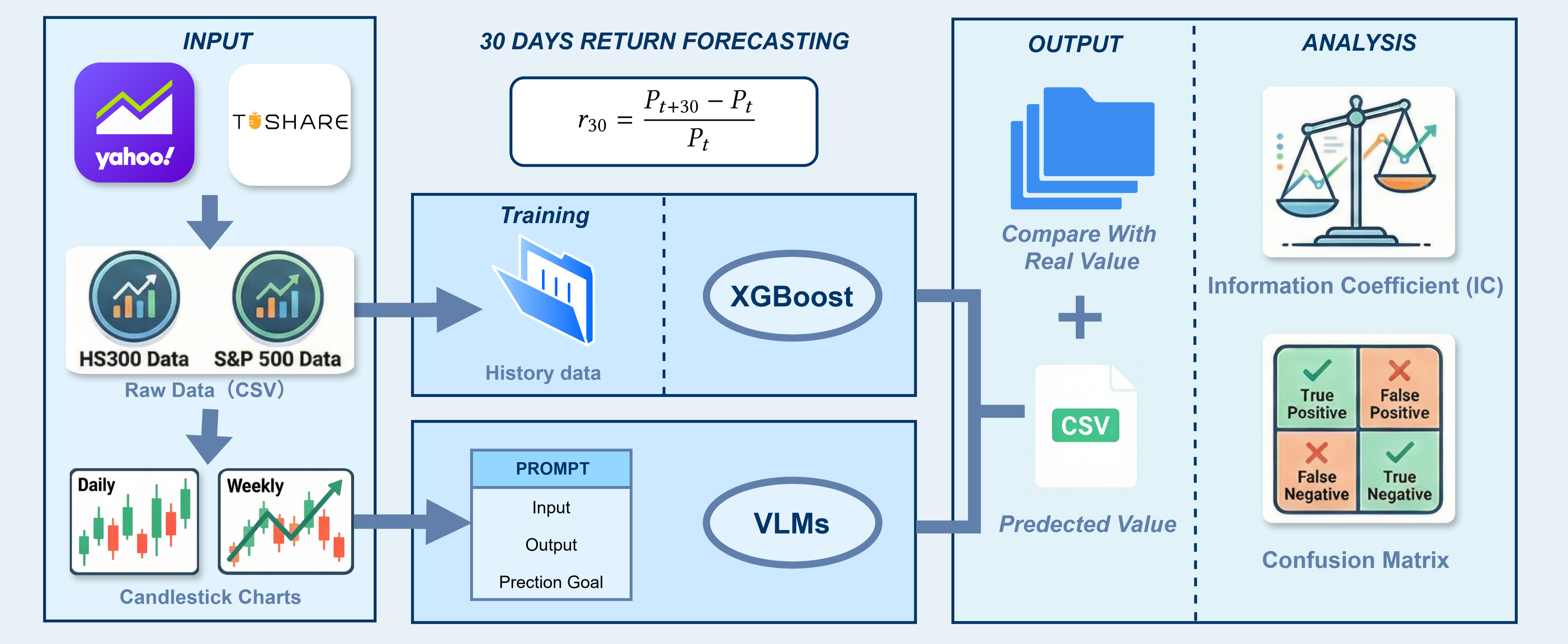}
    \caption{Pipeline for constructing visual inputs and evaluation.}
    \label{fig: 1}
\end{figure*}

\section{Introduction}
With the rapid advancement of artificial intelligence, Vision-Language Models (VLMs) have demonstrated significant potential in financial applications, particularly in stock price forecasting \cite{vista2025vista}. Candlestick charts, as a core analysis tool, visually encapsulate rich market dynamics and price patterns \cite{chootong2012trading}\cite{kusuma2019using}\cite{liang2022stock}\cite{nikam2025stock}, making chart-based prediction a key VLMs' application in finance. However, existing evaluation benchmarks exhibit critical limitations in assessing VLMs' genuine comprehension of stock price information.

Current research faces a fundamental methodological challenge: dataset complexity hinders capability attribution analysis. Mainstream studies employ highly heterogeneous multimodal datasets combining visual and textual content \cite{karadas2025multimodal}\cite{xu2025finmultitime}. While mimicking real trading environments, this introduces critical flaws: unclear capability attribution when achieving strong performance, we cannot determine whether models truly ``understand'' chart patterns or primarily rely on textual signals \cite{vista2025vista}. Most studies lack rigorous ablation experiments to isolate modality contributions, preventing quantification of visual inputs' marginal value.

Furthermore, existing studies neglect the multi-time-scale analysis methodology central to technical analysis \cite{vista2025vista}, thereby limiting the comprehensive assessment of VLMs' capabilities. Professional traders systematically integrate multiple timeframes \cite{khurana2023revolutionize}: long-term frames reveal macro trends forming strategic backdrops \cite{ivanyuk2023long}\cite{khurana2023revolutionize}, while short-term frames provide precise trading signals, breakouts, momentum shifts, and inflection points \cite{roy2022multi}\cite{zhang2025neural}. This hierarchical validation represents a critical cognitive advantage validated through decades of market practice.

To address these limitations, this study proposes a systematic solution minimizing information interference while preserving essential analytical elements. Specifically, our contributions are:

\begin{itemize}
  \item We construct dataset samples that incorporate both daily and weekly candlestick charts with forecast targets, forcing VLMs to extract information purely from visual patterns and eliminating multimodal confounding bias.

  \item We employ a multidimensional evaluation framework that goes beyond traditional information coefficient (IC) metrics by introducing dual-dimensional confusion matrix analysis, which quantifies VLMs' behavioral characteristics across different market states and stock attributes.

  \item We conduct fine-grained analyses examining predictive biases under varying conditions, including false positive and false negative distributions as well as trend capture abilities. This framework provides methodological support for understanding multimodal prediction mechanisms and offers valuable insights for trading strategy optimization.
\end{itemize}

\section{Related Work}
\subsection{Traditional Stock Forecasting Methods}
Stock price forecasting has long been central to financial research. Traditional statistical models, such as ARIMA, demonstrate potential for short-term predictions \cite{khanderwal2021stock} but struggle with complex nonlinear relationships. Machine learning approaches enhance performance through feature engineering: Wu et al. showed XGBoost achieves high accuracy \cite{wu2023stock}, while Sheng et al.'s multimodal LightGBM reached 75.85\% balanced accuracy \cite{sheng2024stock}, though susceptible to overfitting. Deep learning opened new possibilities, with Yu et al. demonstrating ARIMA's trend forecasting utility \cite{yu2024stock} and Nichani et al. validating hybrid ARIMA-LSTM superiority \cite{nichani2024optimizing}. Image-based approaches treat candlestick charts as visual inputs: Sim et al. found CNNs enhance learning through convolution and pooling \cite{sim2019deep}, Chootong et al. integrated neural networks with technical indicators and chart patterns for trading signals \cite{chootong2012trading}, and Kusuma et al. demonstrated CNNs uncover hidden patterns within candlestick images, achieving peak performance across multiple metrics \cite{kusuma2019using}.However, these methods require data or structured image processing and cannot extract effective information directly from candlestick charts for prediction like human traders do.

\subsection{Image Applications and Challenges}
In recent years, financial images have been widely applied to forecasting tasks. However, existing research predominantly employs highly complex multimodal datasets, hindering capability attribution analysis.
Typical studies combine images with multiple information sources: Karadaş et al. incorporated tweet metrics (likes, retweets, comments, follower counts) into feature extraction \cite{karadas2025multimodal}; Xu et al. provided models with pure time series, news sentiment, image trends, and fundamental tables \cite{xu2025finmultitime}; Huang et al. fed FinLLaMA with financial domain knowledge (papers, reports) alongside images, achieving 55.73\% cumulative return and 2.45 Sharpe ratio on TSLA \cite{huang2024open}; Prado et al. visually encoded candlestick charts and identified 16 classic patterns through mathematical rules, finding statistical significance in some patterns \cite{prado2013effectiveness}. While achieving high accuracy, these studies suffer from a fundamental flaw: the inability to determine whether performance stems from visual understanding or textual information. Critically, most lack ablation experiments isolating modality contributions, leaving visual comprehension capabilities a ``black box''.

\subsection{Multi-Time-Scale Analysis Methods}
Multi-time-scale analysis constitutes technical analysis's core methodology. Classical Dow Theory distinguishes primary trends, secondary trends, and short-term fluctuations \cite{edwards2018replacing}; Elliott Wave Theory depicts nested market structures \cite{gartley2012relationship}. Modern systems adopt multi-period confirmation mechanisms like the ``three-screen trading system'' requiring validation across timeframes \cite{forex2012trade5}.
In deep learning, multi-scale modeling has emerged: Xu et al. provided minute, daily, and quarterly resolutions to capture short-term fluctuations, medium-term trends, and long-term fundamentals \cite{xu2025finmultitime}, though primarily targeting tabular data rather than visual candlestick charts. Existing research lacks a systematic evaluation of VLMs' visual integration of multi-period charts. Shu et al. constructed time-series datasets by stitching ten consecutive years of financial charts \cite{shu2025finchart}, yet these experiments failed to analyze whether models truly learned hierarchical reasoning, grasping long-term trends before capturing short-term opportunities, or merely identified stronger statistical correlations in certain periods.

\section{Dataset}
This section describes our process for constructing a multi-scale candlestick chart dataset. The raw market data are obtained from two publicly accessible financial data sources. Historical data for the Chinese market are collected from TuShare, which provides structured daily OHLCV records for A-share stocks in China. Using this interface, we collect data for constituent stocks of HS300 from January 2015 to July 2025, together with trading codes and dates. In addition, historical daily OHLCV records for U.S. stocks are obtained from Yahoo Finance, covering stocks included in the S\&P500 index over the same period.

The multi-period candlestick chart dataset constructed from these historical OHLCV records encompasses diverse market conditions, including bull markets, bear markets, and periods of extreme price volatility, and is used as the visual modality in VLM-related experiments. In contrast, for numerical experiments such as XGBoost, the same historical stock data are processed through a numerical modality construction pipeline to form the numerical modality.

All data access is conducted strictly for academic and noncommercial research purposes and in accordance with the terms of service of the respective data providers. Access to Chinese market data via TuShare is performed using an officially registered account and authenticated API token, while U.S. market data is retrieved from Yahoo Finance through its publicly accessible data interface without authenticated API credentials. The resulting dataset contains only publicly available, market-level financial information and does not involve any personal, private, or sensitive data.

\begin{table}[htbp]
\centering
\caption{Stock Market Regime Distribution (Bull $>$ 70\%, Bear $<$ 30\%)}
\label{tab:market_regime}
\begin{tabular}{lrr}
\toprule
\textbf{Market Regime} & \textbf{Days} & \textbf{Percentage (\%)} \\
\midrule
Bull Market     & 558  & 20.5 \\
Bear Market     & 310  & 11.4 \\
Sideways Market & 1849 & 68.1 \\
\midrule
\textbf{Total}  & 2717 & 100.0 \\
\bottomrule
\end{tabular}
\end{table}

\subsection{Raw Market Data (OHLCV)}
At the raw market data stage, we apply minimal preprocessing to the original OHLCV(open, high, low, close, and volume) time series in order to preserve the underlying market dynamics and avoid introducing model- or modality-specific biases at an early stage. Modality-dependent processing steps for visual and numerical representations are described separately in subsequent subsections.

Based on the closing price series, we define the future return as the target variable of interest, measuring the relative price change over a fixed forward horizon.
Unless otherwise stated, we focus on a 30-day forward return ($H=30$).
The precise mathematical definition is provided in Section~\ref{subsubsec: equation}. 

All raw OHLCV records, together with the corresponding future return values, are aggregated into a unified CSV file, which serves as the common numerical data source for subsequent analysis and evaluation.

\subsection{Visual Modality Data Processing}
To construct the visual modality of our dataset, we generate candlestick chart images at multiple time scales from the raw OHLCV time series using a unified and deterministic pipeline. Each image corresponds to a specific stock, temporal frequency, and cutoff date. The generation process for a single candlestick chart proceeds as follows. 
(1)The configuration parameters and the corresponding stock-level OHLCV data are reconstructed from the input arguments, forming a dedicated dataframe for each stock.
(2)Multi-frequency candlestick data are computed. For the daily frequency, the OHLCV records are sorted by \texttt{trade date}, and moving averages (MA5: black; MA20: blue; MA90: purple) are computed directly from the daily closing prices. For lower-frequency views (weekly), the daily OHLCV series is resampled using standard aggregation rules to obtain open, high, low, close, and volume values for each period, followed by the computation of the same set of moving averages.
(3)The candlestick data corresponding to the specified temporal frequency are selected for visualization.
(4)The time series is truncated according to a predefined cutoff date. For each chart, only historical candlesticks strictly before the specified cutoff date are included, ensuring that no future information is introduced.  
(5)To ensure visual consistency and control input complexity, the number of displayed candlesticks is capped at a maximum of 50 bars. When the available history exceeds this limit, only the most recent 50 candlesticks are retained. 
(6) The candlestick chart is rendered and saved as a PNG image. Moving averages are overlaid using additional plot layers, and trading volume is displayed in a separate lower panel. 

\begin{figure}[H]
    \centering
    \vspace{-10pt}
    \includegraphics[width=1\linewidth]{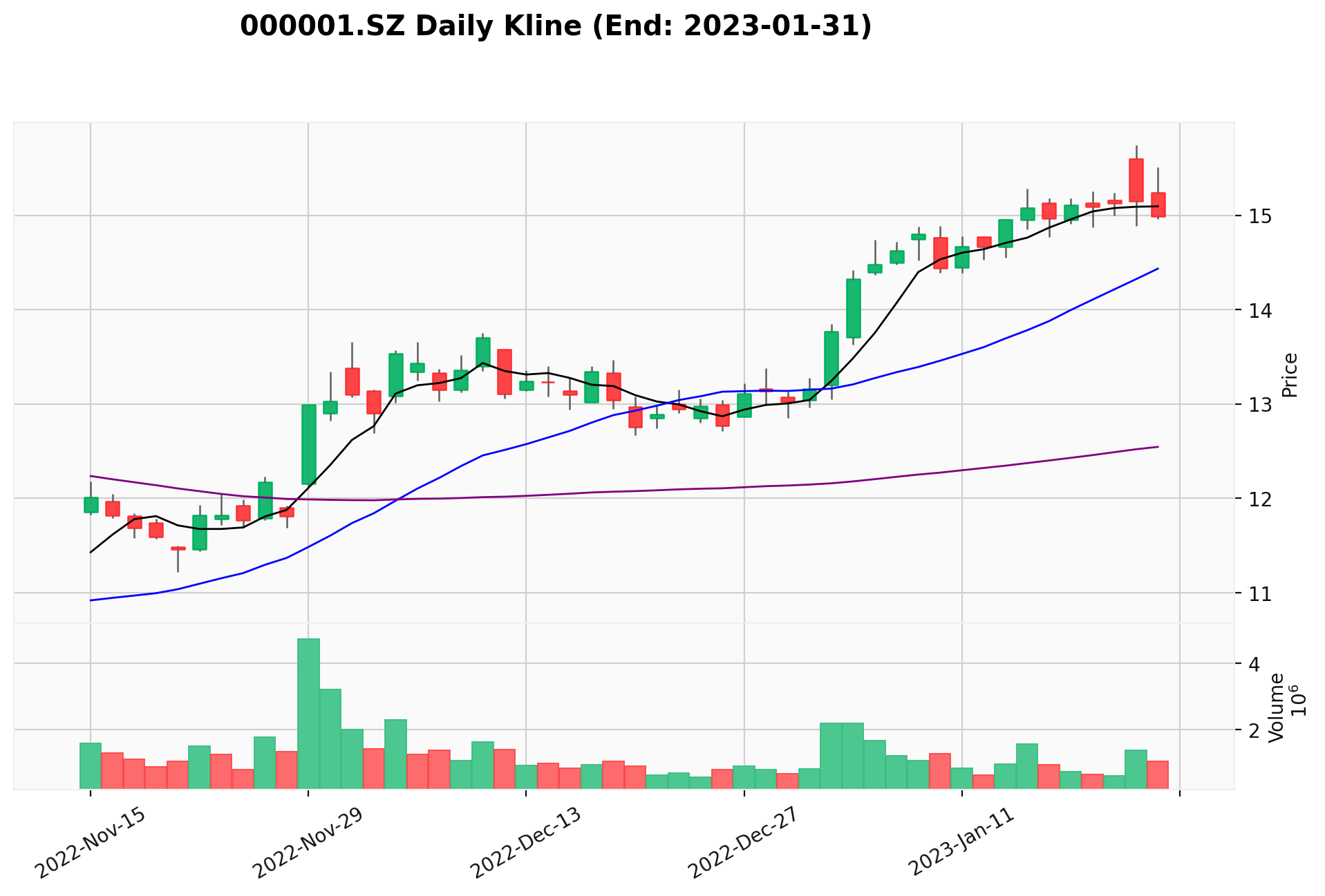}
    \caption{daily candlestick example}
    \label{fig: daily kline}
    \vspace{-10pt}
\end{figure}

\begin{figure}[H]
    \centering
    \includegraphics[width=1\linewidth]{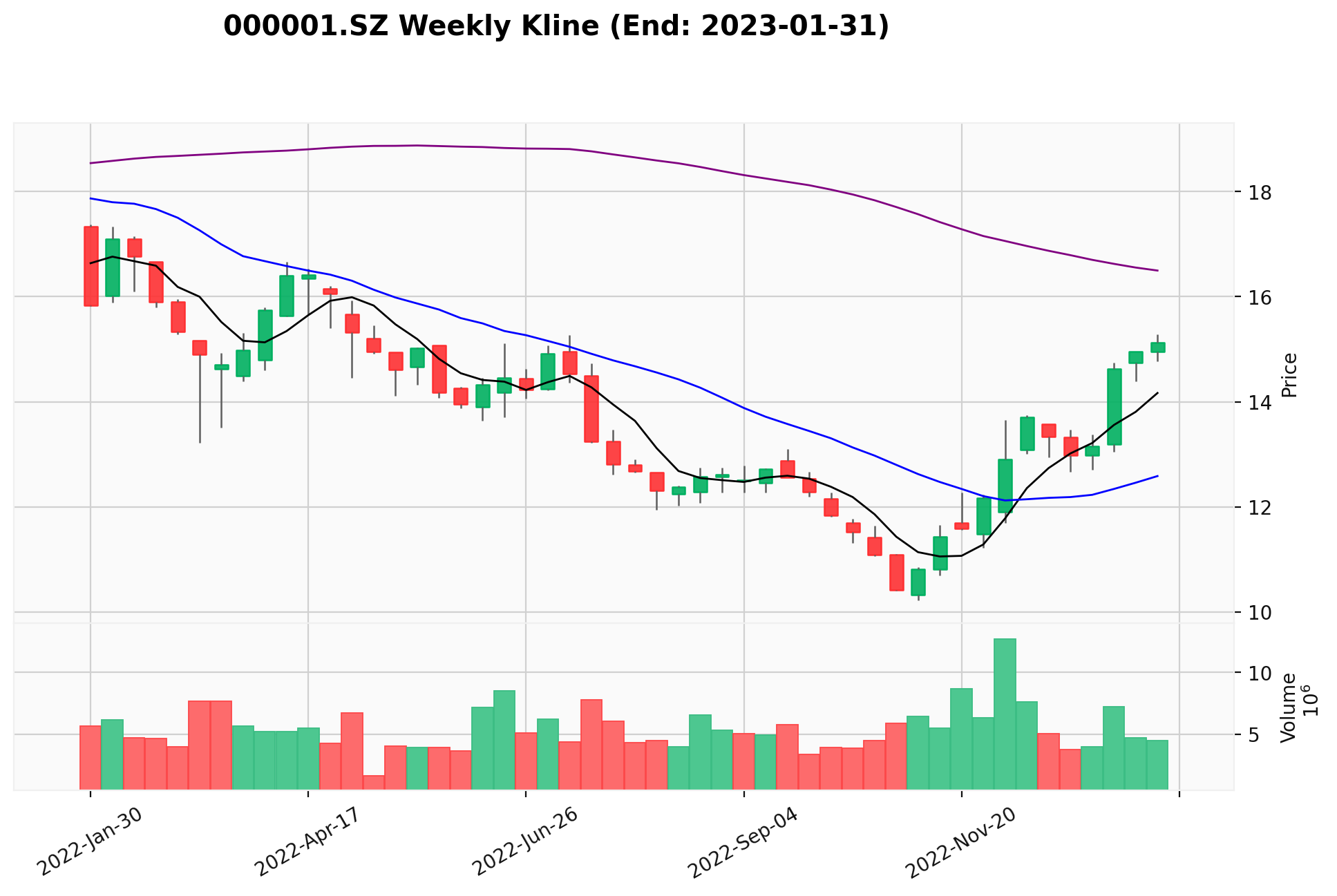}
    \caption{weekly candlestick example}
    \label{fig: weekly kline}
    \vspace{-10pt}
\end{figure}

Brief Summary: Each chart comprises a candlestick price chart with overlaid moving averages and a corresponding volume subplot. The header includes the stock symbol, time frame, and cutoff date, and each chart visualizes up to 50 candlesticks.
Figure\ref{fig: daily kline} and \ref{fig: weekly kline} display the candlestick charts for the same stock at daily and weekly resolutions, respectively. The daily chart shows an upward trend, while the weekly chart indicates a downward trend. The actual 30-day future return for this stock is negative, illustrating how market patterns vary across time scales.

 \vspace{-10pt}
 
After candlestick chart generation, all visual outputs are organized using a hierarchical directory structure to support systematic visual data processing. 
The top-level directory is indexed by the cutoff date associated with each observation, followed by the stock code at the second level. 
Within each stock-specific directory, candlestick charts at different temporal frequencies are stored separately. 
This organization enables the VLMs to access visual representations at multiple time scales for the same asset and cutoff date, facilitating the analysis of price dynamics across multiple time scales in a unified and consistent manner.
To evaluate 30-day return forecasts, candlestick charts are sampled at 30-day intervals as a practical and consistent design choice. 
This sampling strategy simplifies the temporal relationship between visual inputs and target outcomes, improving the clarity and interpretability of the evaluation protocol. 
The 30-day interval is adopted for convenience rather than necessity; alternative sampling strategies are feasible, provided that the sampling interval is aligned with the definition of the future return. 
In contrast, shorter sampling intervals would substantially increase the number of visual inputs and associated API inference costs, while offering limited additional benefit for medium-term forecasting. 
Moreover, charts generated at short intervals tend to be highly similar, differing primarily in recent local price movements, which introduces redundancy and strong temporal correlation among samples.

Overall, this visual data organization and sampling strategy establishes a structured, scalable, and interpretable visual input pipeline, balancing temporal coverage, dataset scale, and sample diversity, and providing a stable foundation for subsequent VLM-based evaluation.

\subsection{Numerical Time-series Data processing}
In addition to the visual modality, numerical time-series inputs are constructed to support baseline models and ensure a fair comparison across modalities.
All numerical features are strictly aligned with the same cutoff dates and future return labels as the visual data, with implementation details deferred to Appendix~\ref{sec:Numerical}.

\subsection{Dataset Statistic}
The historical stock data spans from 2015 to 2025, during which candlestick charts are constructed over the same period. 
We select 300 constituent stocks from the HS300 index and 500 stocks from the S\&P500 index, and generate daily and weekly candlestick charts for 32 cutoff dates.
Table~\ref{tab:total_candlesticks} summarizes the dataset statistics.

\begin{table}[H]
\centering
\caption{Total number of candlestick samples and raw OHLCV records}
\label{tab:total_candlesticks}
\small
\begin{tabular}{lccc|ccc}
\toprule
& \multicolumn{3}{c}{Candlestick Samples} 
& \multicolumn{3}{c}{Raw OHLCV Records (Daily)} \\
\cmidrule(lr){2-4} \cmidrule(lr){5-7}
Frequency 
& HS300 & S\&P500 & Total 
& HS300 & S\&P500 & Total \\
\midrule
Daily  
& 34878 & 61892  & 96770
& 714920 &  1290359 & 2005279 \\

Weekly 
& 34866 & 61888 & 96754
& -- & -- & -- \\

\midrule
Total  
& 69744 & 123780 & 193524
& 714920 &  1290359 & 2005279 \\
\bottomrule
\end{tabular}
\end{table}

\begin{table*}[t]
\centering
\small
\setlength{\tabcolsep}{4pt}
\caption{Comparison with existing datasets.}
\label{tab:dataset_comparison}
\begin{tabular}{l c c c c c c p{3.0cm}}
\toprule
Dataset 
& Modalities 
& Candlestick 
& Multi-scale 
& Visual Reasoning 
& Time Span 
& Market 
& Primary Task 
\\ \midrule
FNSPID (Nasdaq) 
& Text + Price 
& No
& No
& No
& 2009--2023 
& US 
& Frequency forecasting 
\\
FinMultiTime 
& Text + Table + Image + TS 
& Yes
& No 
& No 
& 2009--2025 
& US + China 
& Multimodal forecasting 
\\
Time-MMD 
& Text + Image + TS 
& No
& No 
& No 
& 1989--2024 
& Multi-domain 
& Time-series analysis 
\\
\textbf{Ours} 
& \textbf{Image} 
& \textbf{Yes}
& \textbf{Yes} 
& \textbf{Yes} 
& 2015--2025
& US + China
& \textbf{VLMs testing} 
\\
\bottomrule
\end{tabular}
\end{table*}

\subsection{Comparing to Existing Datasets}
Table~\ref{tab:dataset_comparison} compares our dataset with representative datasets used for time-series or finance forecasting. Existing datasets predominantly focus on textual and numerical signals. For example, FNSPID~\cite{dong2024fnspid} integrates news text with high-frequency price series, while Time-MMD~\cite{liu2024timemmd} provides a general-purpose time-series benchmark with image inputs that are not derived from chart data. Neither includes candlestick charts nor supports reasoning over technical chart structures.
FinMultiTime~\cite{xu2025finmultitime} is the only prior dataset that explicitly incorporates candlestick chart images. However, it treats candlestick charts as auxiliary inputs by compressing long-horizon price movements into coarse trend labels, and therefore does not evaluate models’ native visual understanding of chart structures or cross-scale consistency.

In contrast, our dataset treats candlestick charts as first-class visual inputs and explicitly supports multi-scale candlestick representations. Rather than optimizing for price prediction, our benchmark is designed to evaluate VLMs’ ability to reason over financial charts, including recognizing scale-dependent patterns and maintaining consistency across temporal resolutions. To the best of our knowledge, our dataset is the first to target multi-scale visual reasoning over candlestick charts as a primary evaluation objective.

\section{Benchmark Tasks}

\subsection{Task Definition}
\label{subsec: task}
This study defines the stock trend prediction task as a return regression prediction problem based on visual analysis of candlestick charts.
\subsubsection{Input} 
Each sample contains two candlestick charts for the same stock on the same date: the Daily Chart and the Weekly Chart. This design, with two time scales based on technical analyses across multiple time scales, effectively distinguishes between ``pullbacks within an ongoing trend'' and ``trend reversals'' while testing VLMs’ ability to integrate visual and textual information.
\subsubsection{Output}
Continuous value prediction  \(\hat{r} \in [-0.5,1.0]\) directly corresponds to the expected future return. 
Reasons for choosing regression over classification include: preserving complete granular information for calculating financial metrics like IC; compatibility with classification evaluation (via positive or negative signs); and closer alignment with real investment decision scenarios. The output range [-0.5, 1.0] covers over 99\% of actual cases and permits \(\hat{r}=0.000\) to express ``unclear trend'', avoiding forced predictions. 
\subsubsection{Prediction Goal}
\label{subsubsec: equation}

The 30-day forward yield is calculated as
\begin{equation}
r_{30} = \frac{P_{t+30} - P_t}{P_t},
\label{eq:forward_return}
\end{equation}
where $P_t$ denotes the closing price on the current date, and $P_{t+30}$ represents the closing price 30 trading days later. The 30-day window resides within the ``sweet spot'' of technical analysis validity: it provides sufficient trend continuity while remaining within the applicable range of technical patterns.

\subsection{Prompt Engineering}
To translate the abstract regression prediction task defined in Section~\ref{subsec: task}
into visual analysis instructions executable by VLMs. We designed a structured prompt for models so that they can complete the task in steps. The whole prompt is in the Appendix~\ref{sec: prompt}.

\subsubsection{Role Definition and Task Specification}
The prompt begins by explicitly defining the model's role as ``Stock trend analyst with strong discriminative ability'' to evaluate  ``price fluctuation magnitude'' rather than simple binary judgments, and then emphasizes ``STRICTLY NO bias''. We use two-period candlestick charts as inputs, where green and red candles indicate upward and downward price movements, respectively. The 30-day future return is defined according to Eq.~(\ref{eq:forward_return}).

\subsubsection{Multidimensional Analysis Framework}
This study employs a multidimensional technical analysis framework to predict 30-day stock returns. 
First, candlestick patterns are decomposed to identify body and wick characteristics of red/green candle clusters and capture reversal signals.
Second, the 5-day, 20-day, and 90-day moving average system analyzes price positions, crossovers, and support/resistance effects. 
Third, volume analysis identifies key levels, abnormal spikes, and trend changes. 
Fourth, inflection points detect trend reversals and momentum shifts, with cross-validation against prior indicators. 
Finally, linking daily and weekly signals allows us to examine the consistency between short-term fluctuations and long-term trends.
By integrating price patterns, moving average signals, volume information, and multiple time scales, the framework predicts future returns.

\subsubsection{Output Constraints and Few-Shot Learning}
In this experiment, we enforce a strict output format to standardize model predictions.
The model is required to output a single numerical score in the range [-0.5, 1.0], rounded to three decimal places and enclosed within a <score> tag, without any additional text.
Positive values correspond to bullish predictions, negative values to bearish predictions, and zero indicates uncertainty in trend direction.
To avoid systematic directional bias, the model is instructed to adopt no default preference and to base its predictions solely on the observed technical signals.
Each stock date pair is assigned a unique score.
The provided few-shot examples cover moderate gains, sharp increases, declines, uncertain trends, and extreme cases, with annotations illustrating the correspondence between score magnitude and future returns to guide scale calibration.

\subsection{Evaluation Metrics}
\subsubsection{Confusion Matrices}
\label{subsec: F1 score}
In stock price prediction, TP indicates a predicted rise that actually occurs, TN indicates a predicted fall that actually occurs, FP indicates a predicted rise that does not occur, and FN indicates a predicted fall that does not occur. The TP Rate reflects the ability to capture upward opportunities, while the FP Rate reflects the proportion of false positives.
Accuracy measures general prediction quality; Recall measures the ability to capture upward trends; Precision measures the reliability of predicting increases; Specificity measures the ability to avoid predicting declines; F1 combines prediction reliability and capture capability.
Specific calculation formulas are detailed in Section~\ref{subsubsec: calculation formulas}.

\subsubsection{IC (Information Coefficient)}
The Information Coefficient (IC) measures the linear correlation between predicted values and actual returns. The mean IC reflects the average level of predictive capability, while the median IC indicates predictive stability. The Information Coefficient Ratio (ICIR), calculated as the mean IC divided by the standard deviation of IC, measures risk-adjusted predictive capability. The proportion of statistically significant ICs indicates the frequency of predictive effectiveness.
Rank IC employs Spearman's rank correlation instead of linear correlation, offering greater robustness against outliers. Rank ICIR, mean Rank IC, and statistical significance of Rank IC are similar to those of IC. This metric is suitable for evaluating relative strength judgment capabilities.

\begin{table*}[t]
\centering
\caption{Confusion matrix and performance metrics for different models.}
\label{tab:confusion_merged}
\footnotesize
\setlength{\tabcolsep}{2.5pt}
\begin{tabular}{@{}llrrrrrrr|rrrrr@{}}
\toprule
\textbf{Model} & \textbf{Market} & \textbf{TP} & \textbf{TN} & \textbf{FP} & \textbf{FN} & \textbf{Total} & \textbf{FP Rate (\%)} & \textbf{FN Rate (\%)} & \textbf{Accuracy} & \textbf{Precision} & \textbf{Recall} & \textbf{Specificity} & \textbf{F1} \\
\midrule
Claude-haiku-4-5 & HS300 & 623 & 1834 & 648 & 1670 & 4775 & 13.57 & 34.97 & 51.46\% & 49.02\% & 27.17\% & 73.89\% & 0.3496 \\
Claude-sonnet-4-5 (thinking) & HS300 & 580 & 1170 & 471 & 1051 & 3272 & 14.39 & 32.12 & 53.48\% & 55.19\% & 35.56\% & 71.30\% & 0.4325 \\
\midrule
Gemini2.5-flash (nothinking) & HS300 & 1101 & 1343 & 1139 & 1192 & 4775 & 23.85 & 24.96 & 51.18\% & 49.15\% & 48.02\% & 54.11\% & 0.4858 \\
 & S\&P 500 & 2754 & 1596 & 2324 & 1794 & 8468 & 27.44 & 21.19 & 51.37\% & 54.23\% & 60.55\% & 40.71\% & 0.5722 \\
Gemini2.5-pro (thinking) & HS300 & 1031 & 1473 & 1009 & 1262 & 4775 & 21.13 & 26.43 & 52.44\% & 50.54\% & 44.96\% & 59.35\% & 0.4759 \\
 & S\&P 500 & 2575 & 1739 & 2169 & 1925 & 8408 & 25.80 & 22.89 & 51.31\% & 54.28\% & 57.22\% & 44.50\% & 0.5571 \\
\midrule
GPT4o & HS300 & 978 & 1381 & 1101 & 1315 & 4775 & 23.06 & 27.54 & 49.40\% & 47.04\% & 42.65\% & 55.64\% & 0.4474 \\
GPT5mini & HS300 & 1206 & 1161 & 1321 & 1087 & 4775 & 27.66 & 22.76 & 49.57\% & 47.72\% & 52.59\% & 46.78\% & 0.5004 \\
Qwen & HS300 & 467 & 782 & 562 & 575 & 2386 & 23.55 & 24.10 & 52.35\% & 45.38\% & 44.82\% & 58.18\% & 0.4510 \\
\midrule
XGBoost & HS300 & 775 & 1846 & 840 & 1691 & 5152 & 16.30 & 32.82 & 50.87\% & 47.99\% & 31.43\% & 68.73\% & 0.3798 \\
\bottomrule
\end{tabular}
\end{table*}

\subsection{Peeping Experiment}
This experiment selected two models from the same product line with different release dates: GPT4o-2024-05-13 and GPT4o-2024-11-20. Within the same prediction window (May 14, 2024, to November 19, 2024), both models forecast 30-day future returns based on daily and weekly candlestick charts up to each prediction date, with performance evaluated against actual outcomes. The experimental results show that the IC and several related metrics obtained by the two models are relatively close. No typical peeping pattern, where the later-released model significantly outperforms the earlier version of the model, was observed. Therefore, it is concluded that the selected VLMs did not explicitly utilize actual stock data after its release date to enhance its forecasting performance within this interval, providing preliminary empirical evidence for the interpretable reliability of the prediction results.

\section{Experimental Results}
This study employs a test window spanning from 2023.01.01 to 2025.01.01. Most models are evaluated using the HS300 as the stock universe. Models that have additionally been tested on the S\&P 500 index are marked with parentheses to distinguish their expanded validation scope across different market environments. 

\subsection{Classification Performance Analysis}
\subsubsection{Confusion Matrix}

\begin{figure}[H]
\centering
\includegraphics[width=0.4\textwidth]{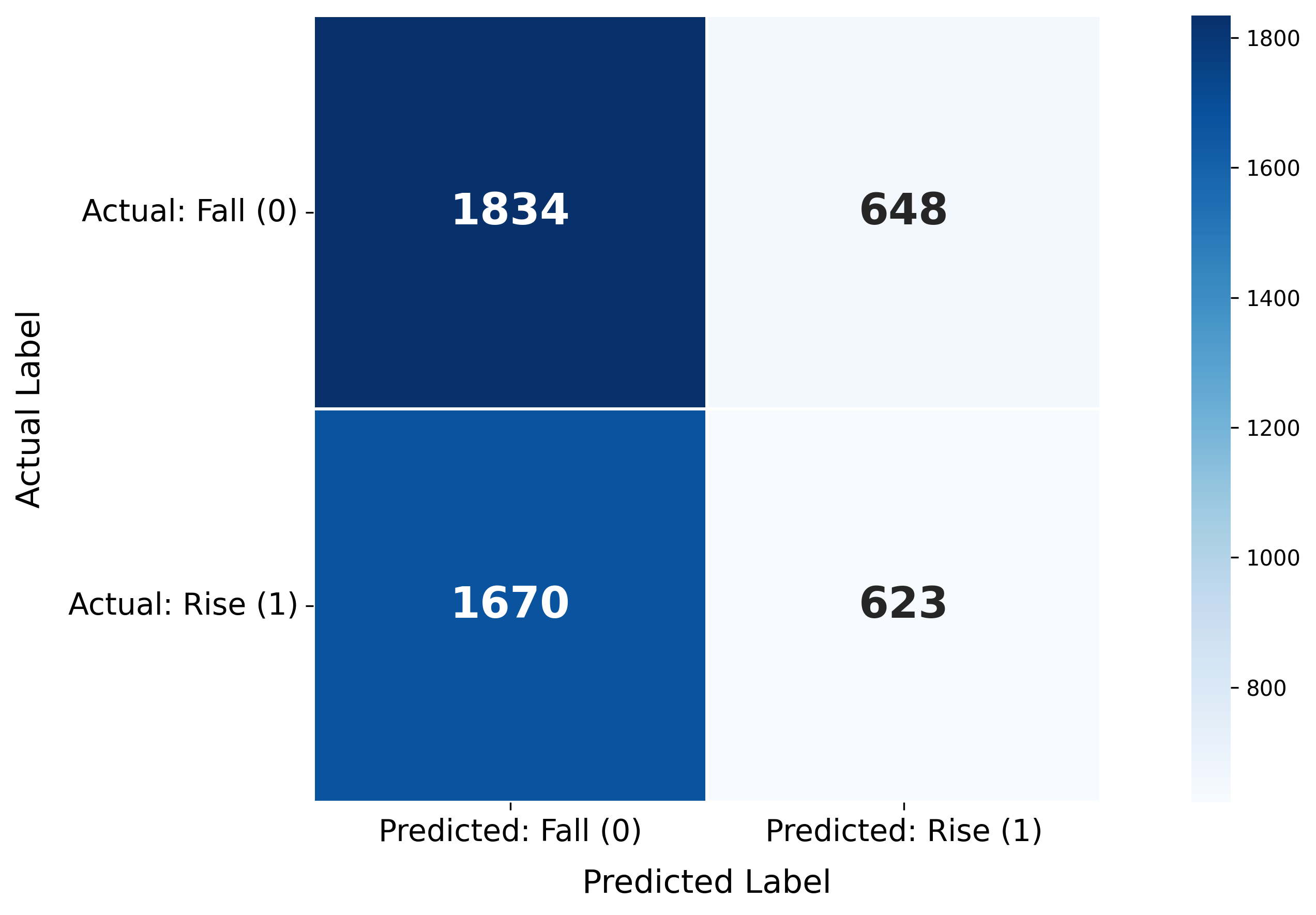}
\caption{Heatmap of Claude-Haiku-4-5}
\label{sonnet}
\end{figure}

The left part of Table ~\ref{tab:confusion_merged} presents the confusion matrix statistics for all models.
The confusion matrix heatmap of the best-performing model, Claude-haiku-4-5, is presented as follows in Figure \ref{sonnet}. 

\vspace{-10pt}

\subsubsection{Metrics based on Confusion Matrix}
\label{subsubsec: calculation formulas}
Based on the confusion matrix, accuracy, precision, recall, specificity, and F1 score can be computed as defined in Section~\ref{subsec: F1 score}. The right part of Table \ref{tab:confusion_merged} presents the performance of various models on these metrics.
\begin{align}
\text{Accuracy} &= \frac{TP + TN}{TP + TN + FP + FN} \\[10pt]
\text{Precision} &= \frac{TP}{TP + FP} \\[10pt]
\text{Recall} &= \frac{TP}{TP + FN} \\[10pt]
\text{Specificity} &= \frac{TN}{TN + FP} \\[10pt]
F_1 &= \frac{2 \times \text{Precision} \times \text{Recall}}{\text{Precision} + \text{Recall}} = \frac{2TP}{2TP + FP + FN}
\end{align}
These findings reveal that different models naturally align with distinct investment philosophies. Claude-Haiku's high precision suits conservative strategies, while GPT5mini's high recall favors aggressive approaches, indicating that model selection should strategically match risk preferences rather than pursuing universal optimization.

\subsection{Prediction Bias Analysis }
\begin{figure}[H]
    \centering
    \includegraphics[width=0.5\textwidth]{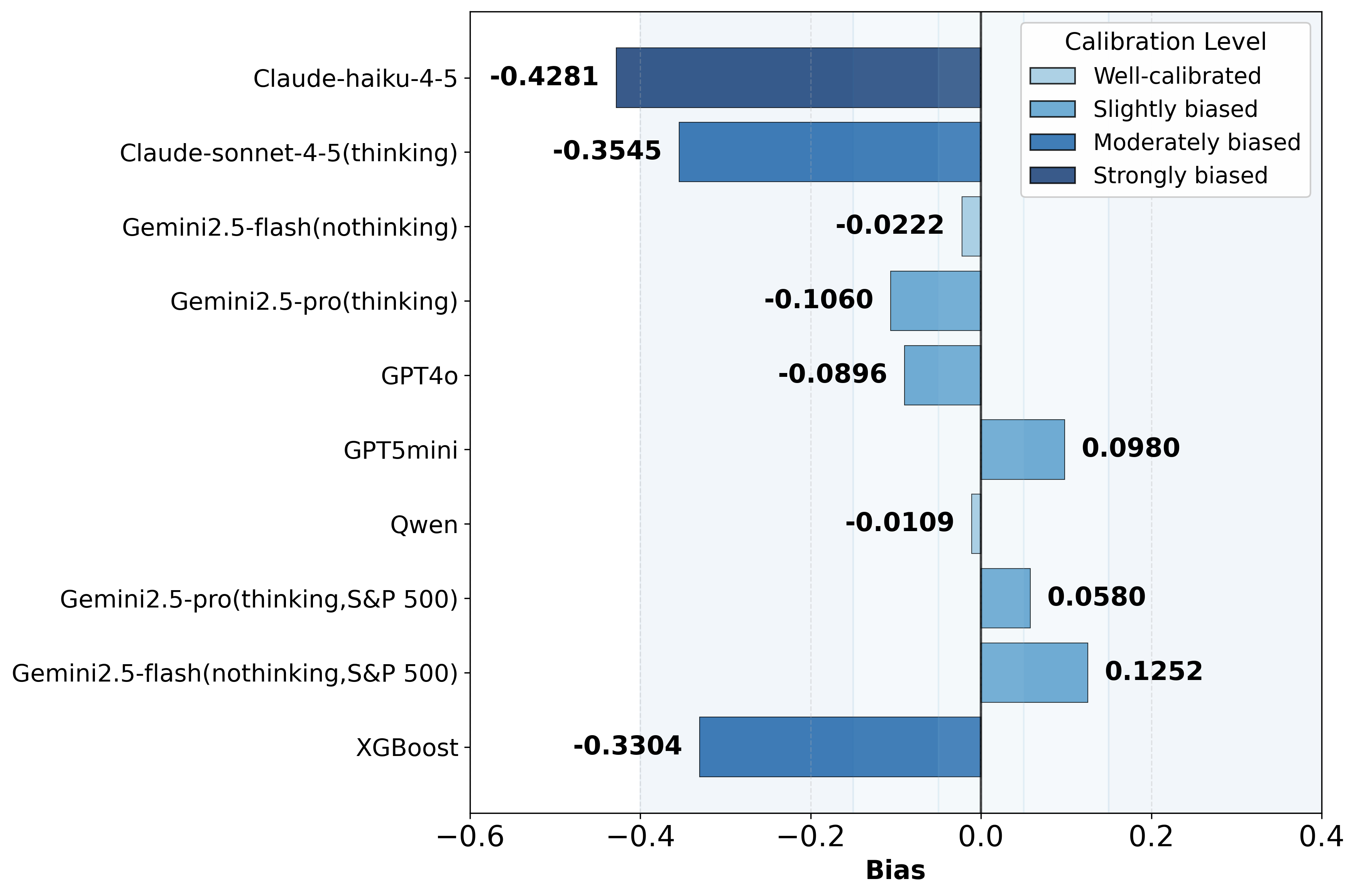}
    \caption{Bias of each model.}
    \label{fig:bias}
    \medskip
    {\footnotesize Well-calibrated ($|\text{bias}| \leq 0.05$), Slightly biased ($0.05 < |\text{bias}| \leq 0.15$),  Moderately biased ($0.15 < |\text{bias}| \leq 0.40$), Strongly biased ($|\text{bias}| > 0.40$).}
\end{figure}

Model calibration evaluates the degree of alignment between the predicted distribution and the true distribution. We employ distribution bias as the calibration metric, defined as:  
\[
\text{Bias} = \left(\frac{N_{\text{pred\_up}}}{N} - \frac{N_{\text{pred\_down}}}{N}\right) - \left(\frac{N_{\text{true\_up}}}{N} - \frac{N_{\text{true\_down}}}{N}\right)
\]
where N denotes the total sample size. A positive bias indicates the model systematically overestimates the probability of upward movements (optimistic), while a negative bias indicates underestimation (pessimistic). Figure~\ref{fig:bias} presents the bias values and calibration grade assessments for each model.

\vspace{-10pt}
\vspace{-10pt}
Model bias evaluation reveals significant divergence across architectures, with traditional Machine Learning and VLMs exhibiting distinct directional preferences. Underlying causes warrant further investigation. This systematic heterogeneity enables constructing ensemble systems: integrating models with complementary biases may achieve bias cancellation and enhanced robustness, offering a promising methodological avenue for financial forecasting by leveraging diverse predictive strengths.

\begin{table*}[t]
\caption{IC Metrics and Rank IC Metrics}
\vspace{-10pt}
\label{tab:ic_rankic_metrics}
{\footnotesize R. denotes rank; Sig. denotes significant ratio; Med. denotes median. Unless otherwise stated, all reported results are obtained from experiments on the HS300 dataset.}
\small
\setlength{\tabcolsep}{3.5pt}
\begin{tabular}{l|rrrrrrrr|rrrrrrrr}
\toprule
\multicolumn{1}{c|}{} &
\multicolumn{8}{c|}{\textbf{IC (Complete Ranking Table)}} &
\multicolumn{8}{c}{\textbf{Rank IC (Complete Ranking Table)}} \\
\cmidrule(lr){2-9} \cmidrule(lr){10-17} 
\textbf{Model} &
\textbf{Mean} &
\textbf{R.} &
\textbf{Med.} &
\textbf{R.} &
\textbf{ICIR} &
\textbf{R.} &
\textbf{Sig.} &
\textbf{R.} &
\textbf{Mean} &
\textbf{R.} &
\textbf{Med.} &
\textbf{R.} &
\textbf{ICIR} &
\textbf{R.} &
\textbf{Sig.} &
\textbf{R.} \\
\midrule
Claude-haiku-4-5 & $-0.01030$ & 9 & 0.03596 & 7 & $-0.04198$ & 9 & 0.813 & \textbf{1} & $-0.00918$ & 8 & 0.01887 & 6 & $-0.03474$ & 8 & 0.750 & \textbf{1} \\
Claude-sonnet-4-5(thinking) & 0.04706 & 2 & 0.05450 & 3 & 0.23602 & 5 & 0.545 & 6 & 0.06198 & 2 & 0.08425 & \textbf{1} & 0.26291 & 4 & 0.636 & 5 \\
Gemini2.5-flash(nothinking) & 0.02576 & 7 & 0.05374 & 4 & 0.09820 & 7 & 0.750 & 3 & 0.02193 & 7 & 0.02210 & 5 & 0.07922 & 7 & 0.688 & 3 \\
Gemini2.5-pro(thinking)& 0.03047 & 5 & 0.07221 & 2 & 0.13092 & 6 & 0.625 & 5 & 0.03603 & 4 & 0.06528 & 2 & 0.13900 & 6 & 0.625 & 6 \\
GPT4o & $-0.01759$ & 10 & 0.04426 & 5 & $-0.06865$ & 10 & 0.750 & 3 & $-0.02024$ & 9 & 0.00526 & 9 & $-0.07286$ & 9 & 0.688 & 3 \\
GPT5mini & $-0.00864$ & 8 & 0.00473 & 9 & $-0.03555$ & 8 & 0.750 & 3 & $-0.02805$ & 10 & $-0.05785$ & 10 & $-0.10490$ & 10 & 0.688 & 3 \\
Qwen & 0.04106 & 3 & 0.03646 & 6 & 0.35515 & 2 & 0.500 & 7 & 0.02844 & 6 & 0.00946 & 8 & 0.19820 & 5 & 0.375 & 8 \\
Gemini2.5-flash(nothinking,S\&P500) & 0.03436 & 4 & $-0.01019$ & 10 & 0.31250 & 3 & 0.412 & 9 & 0.03460 & 5 & 0.01302 & 7 & 0.33200 & 3 & 0.294 & 10 \\
Gemini2.5-pro(thinking,S\&P500) & 0.08990 & \textbf{1} & 0.07628 & \textbf{1} & 0.79370 & \textbf{1} & 0.471 & 8 & 0.07676 & \textbf{1} & 0.06078 & 3 & 0.69660 & \textbf{1} & 0.471 & 7 \\
XGBoost(30d) & 0.03045 & 6 & 0.02951 & 8 & 0.31212 & 4 & 0.250 & 10 & 0.04023 & 3 & 0.02963 & 4 & 0.43373 & 2 & 0.313 & 9 \\
\bottomrule
\end{tabular}
\end{table*}

\subsection{Information Coefficient Analysis}
\begin{figure}[H]
    \centering
    \includegraphics[width=0.5\textwidth]{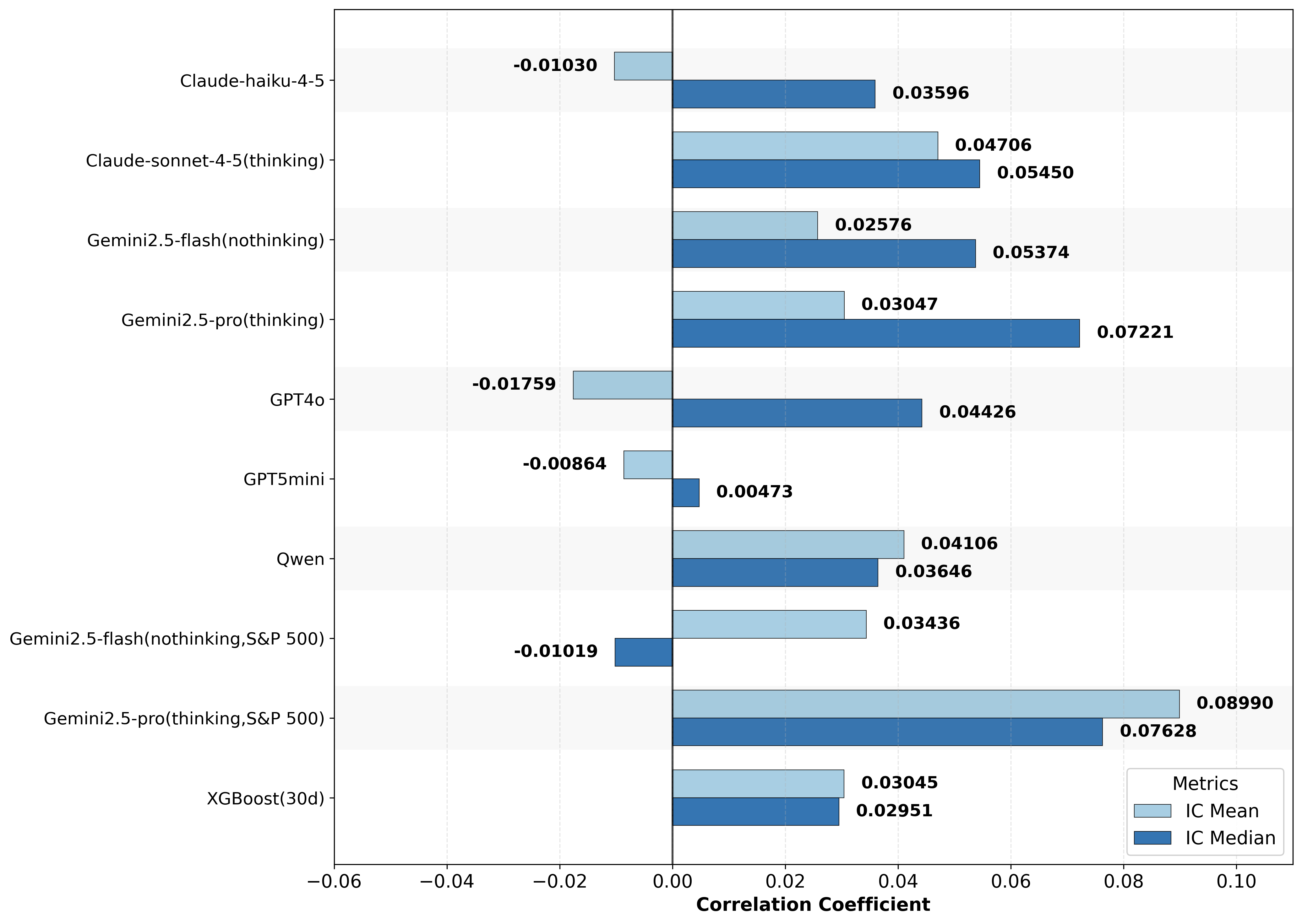}
    \caption{IC Average and IC Median}
    \label{fig:icchart}
\end{figure}

\begin{figure}[H]
    \centering
    \includegraphics[width=0.5\textwidth]{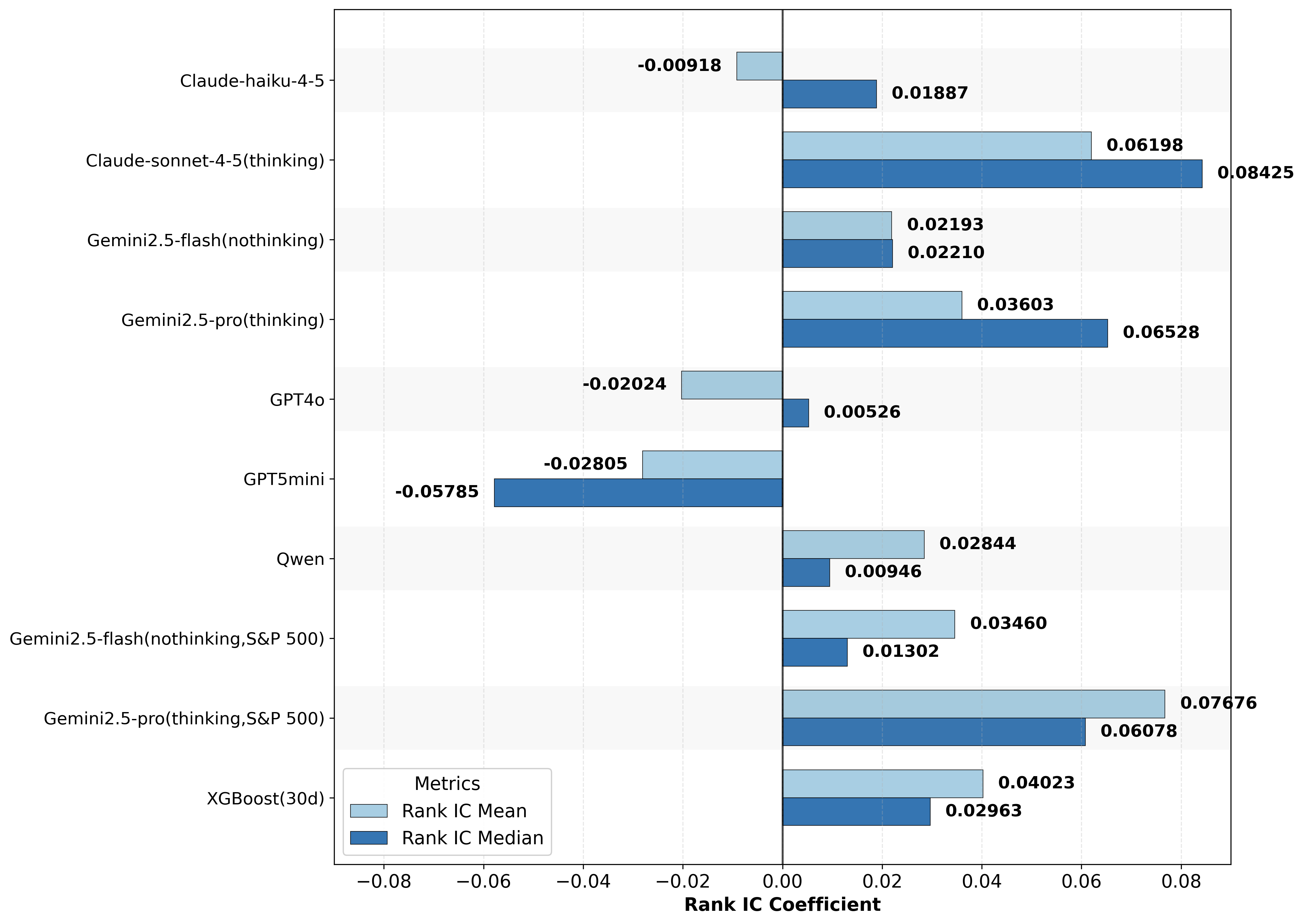}
    \caption{Rank IC Average and Rank IC Median}
    \label{fig:rankicchart}
\end{figure}

The Information Coefficient (IC) measures the correlation between predicted and actual returns, serving as a core metric for evaluating predictive capability. Unlike accuracy, which focuses on directional judgment, IC reflects a model's ability to forecast the magnitude of returns, holding significant importance for portfolio optimization and risk management. Figure~\ref{fig:icchart}, Figure~\ref{fig:rankicchart} and Table~\ref{tab:ic_rankic_metrics}  present the IC and Rank IC series metrics for different models.

\vspace{-10pt}
\vspace{-10pt}
\vspace{-10pt}
\vspace{-10pt}

\subsection{Performance Under Extreme Conditions}

\subsubsection{Bull and Bear Market Performance}
\begin{figure}[H]
    \centering
    \includegraphics[width=0.4\textwidth]{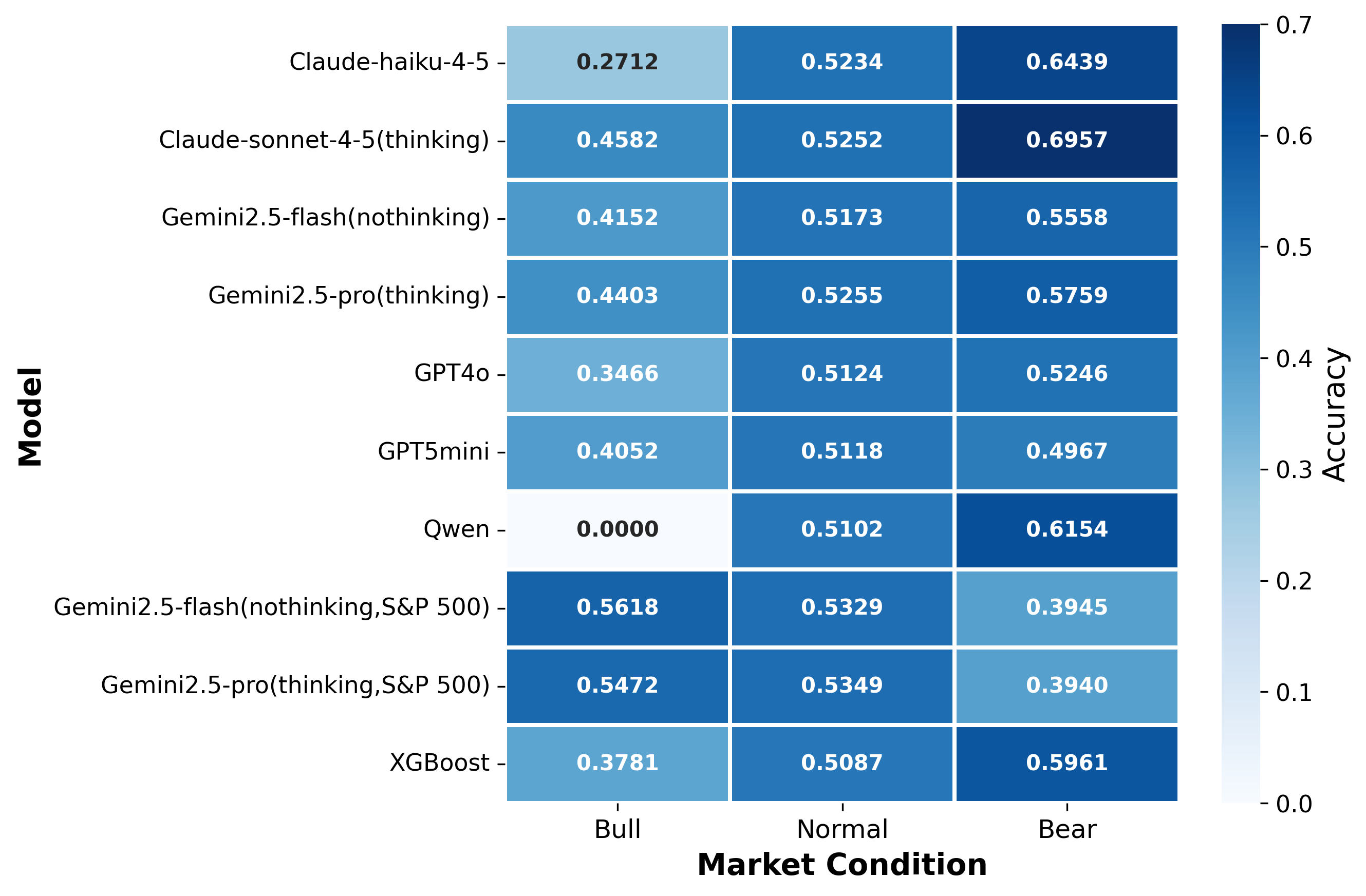}
    \caption{Bull or Bear Market Heatmap}
    \label{fig:ExMarket}
\end{figure}

In this article, market trend is defined as follows: a trading day is deemed as a bull market if the ratio of rising stocks > 70\% on the day, and as a bear market if the ratio of falling stocks > 70\%.Figure \ref{fig:ExMarket} shows each model's performance on these days.
In normal markets, model performance has no significant differences(50.9\%-52.6\%). In extreme markets, divergence emerges. All models demonstrated significantly higher accuracy in bear markets, indicating that models are better at identifying downside risks than capturing upside opportunities. Qwen achieves zero performance because no bull-market instances appear in its test period. This asymmetric predictive capability provides empirical evidence for investors to select models tailored to market conditions.
\vspace{-10pt}
\vspace{-10pt}
\subsubsection{Extreme Stock Cases Analysis}
\begin{figure}[H]
    \centering
    \includegraphics[width=0.4\textwidth]{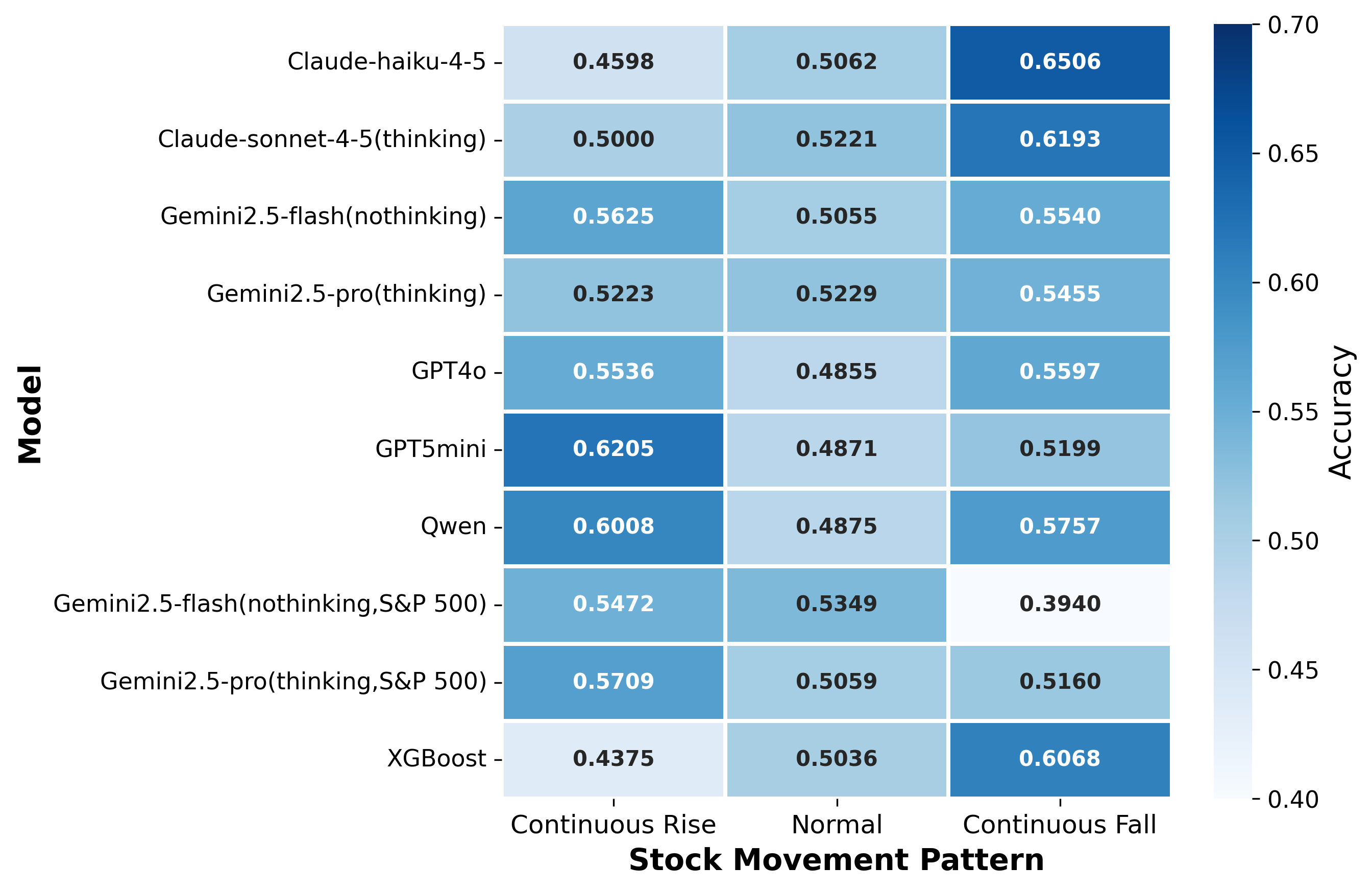}
    \caption{Continuous Rise or Fall Heat-map}
    \label{fig:ExStock}
\end{figure}

In this paper, extreme stocks are defined as follows: an individual stock is continuous rise: if the ratio of the certain stock's rising days > 70\% during the test period, and is continuous fall if the ratio of falling days > 70\%. The predictive behavior of various models toward these stocks is examined, with results presented in the Figure \ref{fig:ExStock}.
Different models exhibit varying accuracy rates for identifying continuously declining stocks and rising stocks, indicating their distinct capabilities in recognizing different stock behaviors. Claude-Haiku demonstrates optimal performance in identifying declining stocks, while GPT5mini excels at capturing upward trends. Notably, XGBoost exhibits the lowest prediction accuracy for rising stocks (43.8\%), indicating its insensitivity to positive momentum signals. This heterogeneity provides a foundation for constructing model portfolio strategies.

\vspace{-10pt}
\vspace{-10pt}
\vspace{-10pt}
\begin{figure}[H]
    \centering
    \includegraphics[width=0.5\textwidth]{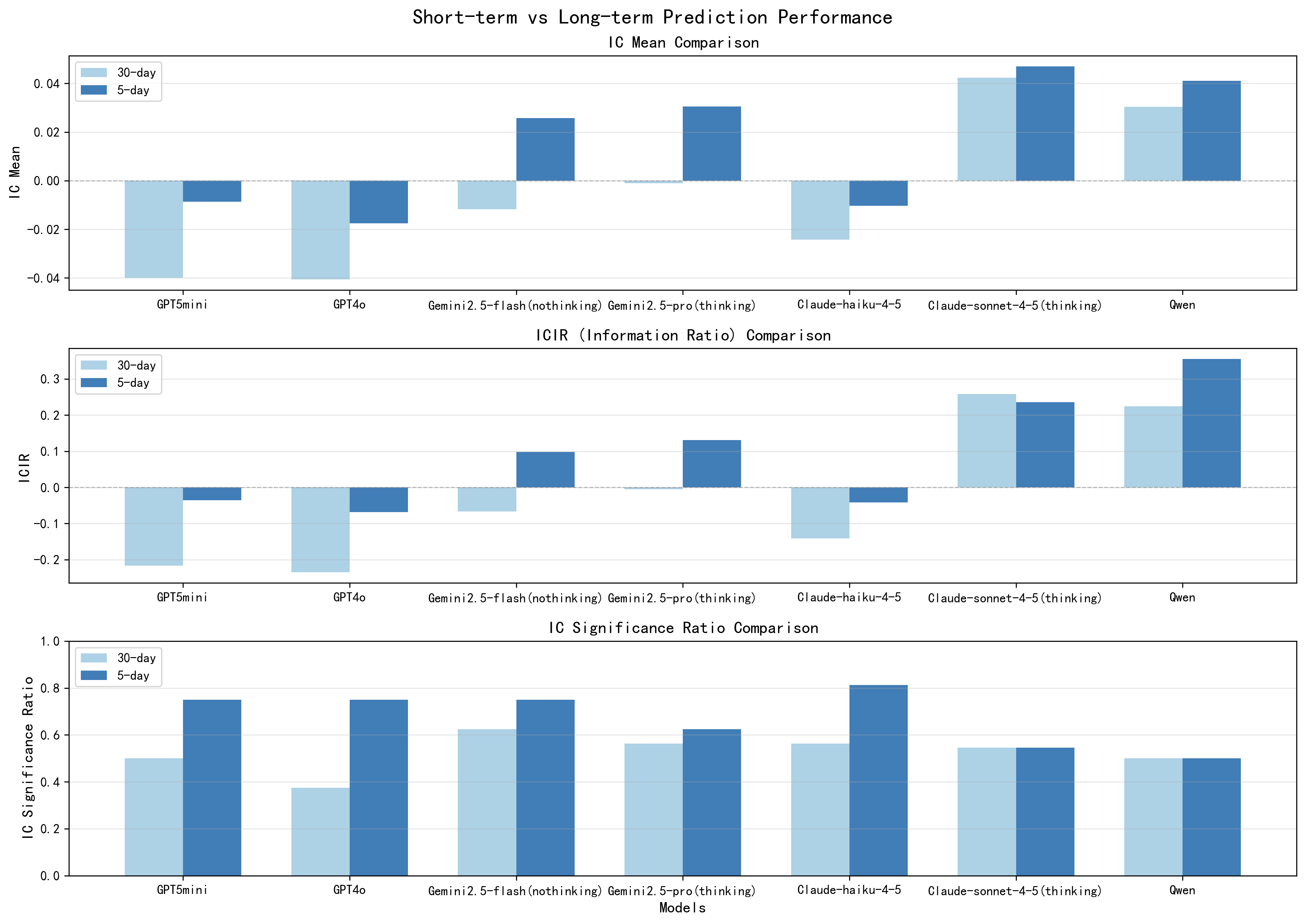}
    \caption{}
    \label{fig:comparison}
\end{figure}
\vspace{-10pt}
\vspace{-10pt}

\subsection{Time Sensitivity Analysis}
Although models were instructed to predict 30-day returns, IC analysis comparing predictions against both 5-day and 30-day actual returns reveals significantly higher correlation with short-term outcomes. This indicates that current VLMs predominantly capture short-term market signals rather than long-horizon trends, exposing fundamental limitations in temporal understanding and task execution for extended prediction horizons. The results are as follows in Figure \ref{fig:comparison}.

IC metric comparison reveals that nearly all models exhibit significantly stronger correlation with 5-day returns than 30-day returns. For instance, Claude-Sonnet-4-5 achieves 5-day IC of 0.047 (ICIR: 0.236) versus 30-day IC of 0.042 (ICIR: 0.258). This discrepancy may stem from VLMs' proficiency in capturing short-term technical patterns abundant in training data, but struggling with long-term predictions requiring deeper fundamental reasoning and managing cumulative uncertainty. These findings suggest VLMs function better as auxiliary tools for short-term trading signals rather than primary support for long-term investment decisions.

\section{Discussion}
Multi-scale candlestick analysis derives value from synchronizing structural fundamentals with dynamic price action across market cycles. However, its inherent latency proves problematic: Technical patterns confirm cycles retrospectively, missing early warning signals of policy shifts or economic inflections, rendering forecasts reactive rather than predictive. Despite this latency, candlestick charts consistently outperform equivalent tabular data, evidencing VLMs' higher performance ceilings. Empirically, VLMs exhibit greater IC/Rank IC significance and systematically elevated median IC (despite comparable means) versus XGBoost, indicating superior predictive stability with reduced tail-risk failures and substantial optimization headroom.

We attribute this to inherent structural privilege in visual representations. Tabular formats encode sequences as atomized numerical vectors, forcing pure statistical discovery of temporal rules. Conversely, candlestick charts externalize domain structure into visual syntax, spatial and morphological patterns encode centuries-refined technical heuristics. This pre-structuring enables VLMs to transfer visual priors from natural images (shape recognition, hierarchical attention) to financial patterns, circumventing sparse-data constraints that tabular models face. The asymmetry reflects differential architectural alignment rather than data superiority: VLMs exploit pre-encoded domain knowledge in visual grammar; tabular models must inductively reconstruct these abstractions from raw features.

\section{Conclusion}
This study aims to construct a multi-period candlestick chart dataset and a standardized evaluation framework to examine the ability of VLMs to utilize multi-scale visual market signals.
This dataset addresses the issues of excessive complexity in existing datasets and insufficient emphasis on multi-period scales. It constructs a minimal dataset in which each sample includes daily and weekly candlestick charts along with the corresponding target value, while maintaining diversity in overall market trends, individual stock behavior patterns, and balanced sample distribution.

The study tested seven widely used commercial VLMs, evaluating them based on model confusion matrices and IC series metrics. Experimental results indicate that while XGBoost, as the baseline, demonstrates superior stability and average ranking capability, certain VLMs show potential in specific areas: Claude-sonnet-4-5(thinking) exhibits strong predictive power in both price direction and return ranking, delivering the best overall performance; Qwen shows balanced direction prediction and strong ranking capability; Gemini2.5-flash achieves the most balanced direction prediction.
We anticipate that future work will incorporate evaluations of additional financial concepts on this dataset, further refine model training, and conduct more detailed investigations into VLMs' comprehension of candlestick charts.

\bibliographystyle{plain}   
\bibliography{ref1}

\appendix

\section{Construction of Numerical Inputs}

\label{sec:Numerical}
Numerical time-series data are constructed to support baseline models and to ensure a fair and controlled comparison with the visual modality. Raw OHLCV records are first validated for completeness and integrity, chronologically ordered, and grouped by stock code to form continuous price series. To match the temporal resolutions used in candlestick chart generation, both daily and weekly OHLCV series are prepared, where the weekly series are obtained through standard aggregation of daily records.

For each prediction task defined by a specific stock and cutoff date, a fixed-length historical window is extracted from the corresponding OHLCV series at the appropriate temporal resolution. These historical segments capture the same market context as the visual inputs and serve as the basis for numerical modeling. The extracted time-series windows are then transformed into numerical feature representations, which are used as inputs to XGBoost models.

Importantly, all numerical inputs are strictly aligned with the same cutoff dates and future return labels as those used in the visual modality. This alignment ensures consistency across different data representations and enables a fair evaluation of numerical baselines against VLM-based models without introducing temporal leakage or information mismatch.

\section{Image Preprocessing and System Optimization}
To balance API cost, inference latency, and visual fidelity, we adopt a standardized image preprocessing and system optimization pipeline for all experiments. All input images are first converted to RGB format. For images with transparent channels, a white background is applied to ensure compatibility with subsequent JPEG encoding. Images are then proportionally resized and resampled using the LANCZOS filter to preserve fine-grained candlestick structures. To reduce bandwidth and API costs while maintaining visual readability, images are compressed using JPEG, yielding approximately a 60\% reduction in API costs in our experiments. Finally, images are encoded in Base64 to conform to the standard HTTP request format.

To improve robustness in parsing model outputs, we employ a three-tier regular-expression–based parsing strategy. The parser first attempts to match the standard output format, \texttt{<score>numeric value</score>}. If a confidence interval is present, the first numerical value is extracted. If neither pattern is matched, the parser falls back to scanning the full response for floating-point values within the valid range of $[-0.5, 1.0]$. Outputs that cannot be reliably parsed are preserved in their original form for subsequent error analysis.
To support large-scale experiments, several system-level optimizations are implemented. Concurrent API requests are limited to five parallel threads, and locking mechanisms are used to prevent duplicate processing. Experimental results are written to CSV files in batches of ten entries, reducing I/O overhead by approximately 90\%. In addition, a resume-from-breakpoint mechanism is enabled by maintaining a set of processed keys, allowing interrupted experiments to be safely resumed. Collectively, these design choices improve the stability, efficiency, and reproducibility of experiments involving thousands of stocks evaluated across multiple models.

\section{Our prompt}
\label{sec: prompt}
\begin{lstlisting}[style=promptstyle]
Stock trend analyst with strong discriminative ability and STRICTLY NO bias: analyze 2 charts (daily/weekly) per stock/date.GREEN = up, RED = down.Predicate:30-day Future Return Rate $=(Future Price - Current Price) / Current Price
Analyze:
1. Decompose charts:
- Red/green candle clusters: body/wick strength, reversal signals
- Moving averages: price position, crossovers, support/resistance
  (5MA = black, 20MA = blue, 90MA = purple)
- Volume: key levels, spikes, and trends
- Inflection points: trend shifts, momentum changes
2. Cross-timeframe link:
- Connect short-term (daily) and long-term (weekly) signals
Rules:
- Output ONLY <score>NUM</score>, where NUM is [-0.5, 1.0] with 3 decimal places.
- +0.001 to +1.000: upward trend
- -0.500 to -0.001: downward trend
- 0.000: unclear trend
- No default positive bias
- One unique score per stock/date
- No additional text
Examples:
<score>0.354</score>
<score>-0.210</score>
<score>0.000</score>
\end{lstlisting}

\section{Limitation and further work}
This study is based solely on evaluation metrics such as confusion matrices and information coefficients for time series, and does not yet include assessment tasks corresponding to financial concepts like resistance levels, support levels, or trend patterns. We anticipate that future work will supplement this dataset with such semantic annotations and evaluations to more accurately assess VLMs' understanding of candlestick chart content. Additionally, the VLMs used in this study are prompt-based and untrained for task-specific purposes, which may limit their predictive performance. Subsequent work will involve further training or fine-tuning of the models to explore VLMs' understanding capabilities of candlestick charts more thoroughly.
Nevertheless, the core significance of this study lies in constructing a candlestick chart dataset with two time scales. We hope our results will inspire further research.

\end{document}